\documentclass{scrartcl}
\usepackage[margin=1.3in]{geometry}
\usepackage{microtype}
\usepackage{graphicx}
\graphicspath{{figures/}}
\usepackage{subfigure}
\usepackage{booktabs}
\usepackage{amsfonts}
\usepackage{amssymb}
\usepackage{amsthm}
\usepackage[parfill]{parskip}
\usepackage{tcolorbox}
\tcbuselibrary{theorems}
\usepackage{amsmath}
\usepackage{hyperref}

\newtcbtheorem[number within=section]{theo}{Theorem}%
{colback=green!3,colframe=green!20!black,sharp corners,fonttitle=\bfseries}{th}

\def\bbR{\mathbb{R}}

\DeclareMathOperator{\Skew}{Skew}

\DeclareMathOperator{\grad}{Grad}

\def\Op{\mathcal{O}_p}
\title{Deep orthogonal linear networks are shallow}
\author{Pierre Ablin \\
CNRS, Département de mathématiques et applications, ENS\\
PSL University}
\date{}
\begin{document}
\maketitle
\begin{abstract}
We consider the problem of training a deep orthogonal linear network, which consists of a product of orthogonal matrices, with no non-linearity in-between.
We show that training the weights with Riemannian gradient descent is equivalent to training the whole factorization by gradient descent.
This means that there is no effect of overparametrization and implicit bias at all in this setting: training such a deep, overparametrized, network is perfectly equivalent to training a one-layer shallow network.
\end{abstract}


\section{Introduction}
How can deep neural networks generalize, when they often have many more parameters than training samples?
The culprit might be the training method, gradient descent, which should be \emph{implicitly biased} towards good local minima that generalize well.

In order to gain better understanding of the dynamics of neural networks trained by gradient descent, one can consider deep linear networks, which are a concatenation of linear transforms, without any non-linearity in between~\cite{saxe2013exact, gunasekar2017implicit, gunasekar2018characterizing,arora2018convergence,ji2018gradient, arora2019implicit}.
This gives an interesting theoretical framework where there is hope to analyse precisely the behavior of gradient descent.

In this work, we consider \emph{deep orthogonal linear networks}, which are deep linear networks where each linear transform is constrained to be orthogonal.
The set of orthogonal matrices is a Riemannian manifold, hence the training is performed with \emph{Riemannian gradient descent}.

We show that training any such network with Riemannian gradient descent is exactly equivalent to training a shallow one-layer neural network, hence fully explaining the role (or lack thereof) of depth in such models.
\section{Background on the orthogonal manifold}

In the following, we let $p$ an integer, and consider square $p\times p$ matrices.
A matrix $W \in \bbR^{p\times p}$ is orthogonal when $WW^{\top} = I_p$.
The set of such matrices is called the orthogonal manifold and is denoted $\Op$~\cite{edelman1998geometry}.
This set is a Riemannian manifold: around each point $ W$, it locally resembles a flat Euclidean space called the tangent space at $W$, denoted $T_W$.
The tangent space has a simple structure: it is the set of matrices that write $AW$ for $A\in \Skew_p$ a skew-symmetric matrix.
We can then compute the Euclidean projection of a matrix $B$ to the tangent space by $\mathcal{P}_{T_W}(B) = \Skew(BW^{\top})W$, where the $\Skew$ of a matrix $M$ is $\Skew(M) = \frac12(M-M^{\top})$, its skew-symmetric part.

Next, we consider the task of minimizing a differentiable function $\ell:\bbR^{p\times p} \to \bbR$ on the manifold $\Op$.
The Riemannian gradient of $\ell$ at $W\in \Op$ is the projection of the Euclidean gradient of $f$ on $T_W$, and is therefore:
$$
\grad \ell(W) = \Skew(\nabla \ell(W)W^{\top})W\enspace.
$$
In order to obtain a working optimization algorithm, and move in the opposite direction of the gradient while staying on the manifold, one should use a \emph{retraction} $\mathcal{R}$ which takes as input an orthogonal matrix $W$ and a vector $X\in T_B$, such that $\mathcal{R}(W, X) \in \Op$ and, as $X\to 0$:
$$
\mathcal{R}(W, X) =W + X + O(\|X\|^2)\enspace.
$$
The Riemannian gradient descent with step size $\eta> 0$ then iterates~\cite{absil2009optimization}
$$
W\leftarrow \mathcal{R}(W, -\eta \grad \ell(W))\enspace.
$$

For concreteness, in the remainder of this note, we consider that the retraction is the exponential, namely:

$$
\mathcal{R}(W, X) = \exp(XW^{\top})W.
$$

As a consequence, Riemannian gradient descent iterates:
\begin{equation}
    \label{eq:rgd}
    W\leftarrow \exp\left(-\eta \Skew(\nabla \ell(W) W^{\top})\right)W.
\end{equation}

One could also consider different retractions, like the Cayley transform or the projection, which would not change the results below.
We are now ready to turn our attention to deep orthogonal matrix factorization.
\section{Deep orthogonal matrix factorization}
In the following, we consider a deep linear orthogonal network of depth $L$, parametrized by the $L$ orthogonal matrices $W_1, \dots, W_L \in \Op$.
This network maps inputs $x_1\in\bbR^p$ to outputs $x_{L + 1}\in\bbR^p$ with the recursion
$$
\text{For } i=1\dots L,\enspace x_{i+1} = W_i x_i\enspace.
$$
Since this network is linear, we have the trivial relationship $x_{L+1} = \Pi x_1$, where $\Pi = W_L\times \dots \times W_1$.
We therefore identify the network and the linear orthogonal transform $\Pi$.

The network is then trained by minimizing a cost function.
We let a differentiable function $\ell : \bbR^{p\times p}\to \bbR$, and denote by $G(X)\in \bbR^{p\times p}$ its gradient.
The network is trained by finding weights $W_1,\cdots, W_L$ that satisfy the optimization problem:
\begin{equation}
    \label{eq:deep_ortho_general}
    \min_{W_1,\dots, W_L \in \Op} \mathcal{L}(W_1, \dots, W_L) \triangleq\frac1L \ell(W_L\times \dots\times W_1)  
\end{equation}

The rest of this note is devoted to the study of the trajectory of Riemannian gradient descent for this problem.
We start by computing the derivatives of $\mathcal{L}$
\paragraph{Derivatives}
The Euclidean gradient of $\mathcal{L}$ with respect to one of the $W_i$ is 
\begin{align}
\nabla_i\mathcal{L} &=\frac1LW_{i+1}^{\top}\times \dots \times W_L^{\top}\times G(W_L \times\dots\times W_1)\times W_1^{\top}\times\dots\times W_{i-1}^{\top}\\
\label{eq:gradient}
&=\frac1L \Pi_{i}^{\top} G(\Pi) \Pi^{\top}\Pi_iW_i\enspace, 
\end{align}

where $\Pi_{i} = W_L\times \dots \times W_{i+1}$ and $\Pi = W_L\times \dots \times W_1$.
These product matrices are orthogonal, since all the $W_i$ are assumed orthogonal.
The Riemannian gradient of $\mathcal{L}$ with respect to $W_i$ is the projection of $\nabla_i\mathcal{L}$ on the tangent space of $\Op$ at $W_i$, and simple computations give
$$
\grad_i\mathcal{L} = \psi_i W_i\enspace, 
$$
where $\psi_i = \frac1L\Skew((\nabla_i\mathcal{L}) W_i^{\top})$ is a skew-symmetric matrix.
Following the formula~\eqref{eq:gradient}, we find
\begin{align}
    \label{eq:relat_grad}
    \psi_i&=\frac1L \Skew(\Pi_i^{\top} G(\Pi)\Pi^{\top}\Pi_i)\\
    \label{eq:relat_grad2}
    &=\frac1L \Pi_i^{\top} \Skew(G(\Pi)\Pi^{\top}) \Pi_i,
\end{align}
where for~\eqref{eq:relat_grad2} we used the equivariance of $\Skew$ with respect to an orthogonal change of basis.

\paragraph{Optimization}
We consider a simultaneous update on the $W_i$, using the Riemannian gradient descent on $\mathcal{L}$.
One step of gradient descent with the exponential retraction transforms the $W_i$ into $W'_i$ with the formula:
\begin{equation}
    \label{eq:update}
    W_i' = \exp(-\eta\psi_i) W_i\enspace,
\end{equation}
with $\eta >0$ a step size.
Using~\autoref{eq:relat_grad2} and the equivariance of $\exp$, we find $W_i' = \Pi_i^{\top}\exp\left(-\frac\eta n\Skew(G(\Pi) \Pi^{\top})\right)\Pi_{i-1}$.

It is then easily seen that the product $\Pi' = W_L'\times\dots\times W'_1$ simplifies to give:
\begin{equation}
    \label{eq:product_next}
    \boxed{\Pi' = \exp\left(-\eta\Skew(G(\Pi) \Pi^{\top})\right) \Pi}\enspace
\end{equation}
The similarity with~\autoref{eq:rgd} is striking: this is exactly the update equation of Riemannian gradient descent on $\Pi$ when one directly minimizes $\ell(\Pi)$.
This shows that the network architecture has absolutely no impact on the trajectory of the overall learned transform, apart through initialization: \textbf{training any deep orthogonal matrix factorization network with gradient descent is equivalent to training a shallow one-layer network}.
\paragraph{Continuous setting}
Letting the step $\eta$ be infinitesimal in \autoref{eq:product_next} gives the flow equation 
$$
\frac{d\Pi}{dt} = - \Skew(G(\Pi)\Pi^{\top})\Pi
$$
which unsurprisingly corresponds to the Riemannian gradient flow equation for the minimization of $\ell(\Pi)$: the same conclusion holds in the continuous training setting.
\paragraph{Dependence on initialization only}
The trajectory of the product $\Pi$ across iterations verifies the recursion of usual gradient descent, hence it only depends on its initial values.
As a consequence, two networks of different depths initialized with weights $(W_1, \dots,W_L)$ and $(U_1, \dots, U_{L'})$ such that $W_L\times\dots\times W_1 =U_{L'}\times\dots\times U_1$ will always represent the same mapping if they are trained with gradient descent with same steps.

\section*{Conclusion}
We have shown that training a deep linear orthogonal network is equivalent to training a one layer neural network.
As a consequence, depth or overparametrization play no role in this setting: adding more layers to the network cannot change the trajectory of the mapping during gradient descent. 

\subsection*{Acknowledgement}
This work was supported* by the French government under management of Agence Nationale de la Recherche as part of the “Investissements d'avenir” program, references ANR19-P3IA-0001 (PRAIRIE 3IA Institute).
\bibliography{biblio}

\begin{thebibliography}{1}

\bibitem{absil2009optimization}
P.-A. Absil, R.~Mahony, and R.~Sepulchre.
\newblock {\em Optimization algorithms on matrix manifolds}.
\newblock Princeton University Press, 2009.

\bibitem{arora2018convergence}
S.~Arora, N.~Cohen, N.~Golowich, and W.~Hu.
\newblock A convergence analysis of gradient descent for deep linear neural
  networks.
\newblock {\em arXiv preprint arXiv:1810.02281}, 2018.

\bibitem{arora2019implicit}
S.~Arora, N.~Cohen, W.~Hu, and Y.~Luo.
\newblock Implicit regularization in deep matrix factorization.
\newblock In {\em Advances in Neural Information Processing Systems}, pages
  7413--7424, 2019.

\bibitem{edelman1998geometry}
A.~Edelman, T.~A. Arias, and S.~T. Smith.
\newblock The geometry of algorithms with orthogonality constraints.
\newblock {\em SIAM journal on Matrix Analysis and Applications},
  20(2):303--353, 1998.

\bibitem{gunasekar2018characterizing}
S.~Gunasekar, J.~Lee, D.~Soudry, and N.~Srebro.
\newblock Characterizing implicit bias in terms of optimization geometry.
\newblock {\em arXiv preprint arXiv:1802.08246}, 2018.

\bibitem{gunasekar2017implicit}
S.~Gunasekar, B.~E. Woodworth, S.~Bhojanapalli, B.~Neyshabur, and N.~Srebro.
\newblock Implicit regularization in matrix factorization.
\newblock In {\em Advances in Neural Information Processing Systems}, pages
  6151--6159, 2017.

\bibitem{ji2018gradient}
Z.~Ji and M.~Telgarsky.
\newblock Gradient descent aligns the layers of deep linear networks.
\newblock {\em arXiv preprint arXiv:1810.02032}, 2018.

\bibitem{saxe2013exact}
A.~M. Saxe, J.~L. McClelland, and S.~Ganguli.
\newblock Exact solutions to the nonlinear dynamics of learning in deep linear
  neural networks.
\newblock {\em arXiv preprint arXiv:1312.6120}, 2013.

\end{thebibliography}
\bibliographystyle{abbrv}
\end{document}